\documentclass[10pt,twocolumn,letterpaper]{article}

\usepackage{cvpr}              

\usepackage[dvipsnames]{xcolor}
\usepackage{array} 

\definecolor{cvprblue}{rgb}{0.21,0.49,0.74}
\usepackage[pagebackref,breaklinks,colorlinks,citecolor=cvprblue]{hyperref}

\def\paperID{37} 
\def\confName{CVPR}
\def\confYear{2024}

\title{AnimalMotionCLIP: Embedding motion in CLIP for Animal Behavior Analysis}

\author{
Enmin Zhong  \; \; Carlos R. del-Blanco  \; \; Daniel Berjón  \;\; Fernando Jaureguizar \;\;  Narciso Garc\'ia \\
Grupo de Tratamiento de Im\'agenes (GTI), Information Processing and Telecommunications Center, \\
ETSI~Telecomunicaci\'on, Universidad Polit\'ecnica de Madrid \\
{\tt\small \{enmin.zhong, carlosrob.delblanco, daniel.berjon, fernando.jaureguizar, narciso.garcia\}@upm.es }
}

\begin{document}
\maketitle
\begin{abstract}

Recently, there has been a surge of interest in applying deep learning techniques to animal behavior recognition, particularly leveraging pre-trained visual language models, such as CLIP, due to their remarkable generalization capacity across various downstream tasks. However, adapting these models to the specific domain of animal behavior recognition presents two significant challenges: integrating motion information and devising an effective temporal modeling scheme. In this paper, we propose AnimalMotionCLIP to address these challenges by interleaving video frames and optical flow information in the CLIP framework. Additionally, several temporal modeling schemes using an aggregation of classifiers are proposed and compared: dense, semi-dense, and sparse. As a result, fine temporal actions can be correctly recognized, which is of vital importance in animal behavior analysis. Experiments on the Animal Kingdom dataset demonstrate that AnimalMotionCLIP achieves superior performance compared to state-of-the-art approaches. 

\end{abstract}    
\section{Introduction}
\label{sec:intro}

Understanding animal behavior is crucial for various fields, such as wildlife conservation, ecology, and animal welfare. However, automatically recognizing animal behavior presents a significant challenge due to the vast diversity in size, shape, and appearance among different species and even animals of the same species. In addition, environmental conditions, including different backgrounds and habitats, add another layer of complexity to the task. 

Current automatic animal behavior recognition approaches predominantly rely on video deep-learning techniques \cite{feichtenhofer2019slowfast, feichtenhofer2020x3d,carreira2018quo}, which are trained on a predefined and closed set of behaviors (classes). These behaviors are usually limited to specific species in controlled environments, such as mice in laboratories, sheep and cows in the livestock sector \cite{CHENG2022107010}, or salmon in aquaculture \cite{MALOY2019105087}. However, real-world scenarios present animals in diverse, complex, and cluttered environments, interacting with each other or their surroundings. To face these challenges, vast amounts of new annotated data from these real-world scenarios are typically necessary to capture these nuances, due to the difficulty in transferring the learned behaviors among different animals and to novel behavior categories. Consequently, this strategy is costly and not scalable with the vast range of animal species. 
Existing datasets for animal behavior recognition are notably scarce and often suffer from limitations similar to those encountered in animal action recognition methods. These datasets, such as the MammalNet \cite{chen2023mammalnet} and the KABR dataset \cite{Kholiavchenko2024KABRID}, are constrained by their limited diversity. Specifically, the KABR dataset contains videos of three animal categories exhibiting seven types of behaviors, while MammalNet includes videos covering twelve behavior categories of mammals. This restricted range of species and behaviors limits the variety of data available for effective training and evaluation. The large-scale AnimalKingdom dataset \cite{ng2022animal} addresses these limitations. It comprises 850 species of mammals, fishes, amphibians, reptiles, birds, and insects, with 140 action classes spanning life stages, daily activities, and social interactions (e.g., molting, feeding, playing). 

Recently, a new paradigm in transfer learning has appeared, based on large pre-trained visual language models like CLIP (Contrastive Language-Image Pre-training) \cite{radford2021learning}, which are able to learn general representations based on paired web-scale text-image datasets. They have shown great generalizability and transferability to downstream tasks, especially in human action recognition tasks \cite{ju2022prompting,ni2022expanding,wasim2023vitaclip,rasheed2023finetuned}. 
However, its application in animal behavior analysis remains relatively scarce compared to their use in human action recognition. Current studies that apply cross-modality models (text-image) in animal behavior recognition often adopt design principles similar to those used in ActionCLIP \cite{wang2021actionclip} or Vita-CLIP \cite{wasim2023vitaclip} for human action recognition. These methods typically apply prompt-based learning, in which specific prompt vectors are designed for the text encoder input while keeping the entire model frozen. Although this prompting-based method enhances the model's performance, the results are often difficult to interpret and tend to be affected by noisy labels \cite{Zhou_2022}. For example, Jing et al. \cite{10.1145/3581783.3612551} introduced a category-specific prompting technique, extending the CLIP model with an additional branch that predicts animal classes. However, this animal category specialization may be biased towards specific animal classes and can hinder the model's ability to accurately recognize behaviors from unseen or underrepresented classes. 
Mondal et al. \cite{Mondal_2023} proposed MSQNet (Multimodal Semantic Query Network) for general action recognition in humans and animals. Although merging vision and language information leads to better performance in animal behavior recognition, the approach adopted in MSQNet requires a computationally intensive pre-training on the massive Kinetics-400 dataset \cite{kay2017kinetics}, followed by fine-tuning on the target Animal Kingdom dataset \cite{ng2022animal} for the prediction task.
The recent WildCLIP \cite{Gabeff2024} proposed a combination of prompting-based techniques and a feature adapter for CLIP to generate annotations for images of wild animals. Different from WildCLIP, our proposal focuses on animal behavior recognition at the video level. Video-level action recognition involves understanding the temporal dynamics and sequences of movements across multiple frames, making it a more complex and computationally intensive task. 

Another common limitation of previous works is that they tend to overlook motion information and the temporal modeling scheme. These challenges are even more notorious in the animal domain, since subtle motion and temporal information can lead to different behaviors.
Regarding motion information, changes in velocity, acceleration, or movement patterns can convey important behavioral cues to distinguish different behaviors, such as aggression, exploration, or mating behavior. Optical flow can obtain the required fine dynamic information from a video, capturing subtle nuances in animal movements that may not be evident from still frames alone. On the other hand, temporal resolution is also fundamental in distinguishing different behaviors since an action can occur in seconds, minutes (or even hours). 

To address the previous limitations, a visual language model that explicitly includes motion information is proposed. In addition, we evaluate different temporal sampling schemes to identify the most effective approach.
First, dense optical flow information is interleaved with color frames at the frame level to embed fine motion information in the pre-trained visual-language model of CLIP. Then, several temporal resolution models are proposed and compared using a combination of classifiers. Every classifier focuses on a different sampled selection of frames following a dense, semi-dense, and sparse sampling scheme to take into account a diverse set of global behavior contexts. Thus, every classifier yields a set of partial inference scores, which are finally aggregated to predict the behavior occurring in the video. Experimental results show that our system has a remarkable generalization capacity for animal behavior recognition and performs better than the state-of-the-art action recognition algorithms in the Animal Kingdom dataset.

\section{Proposed method} 
\label{sec:method}

\begin{figure}
    \centering
    \includegraphics[width=0.99\linewidth]{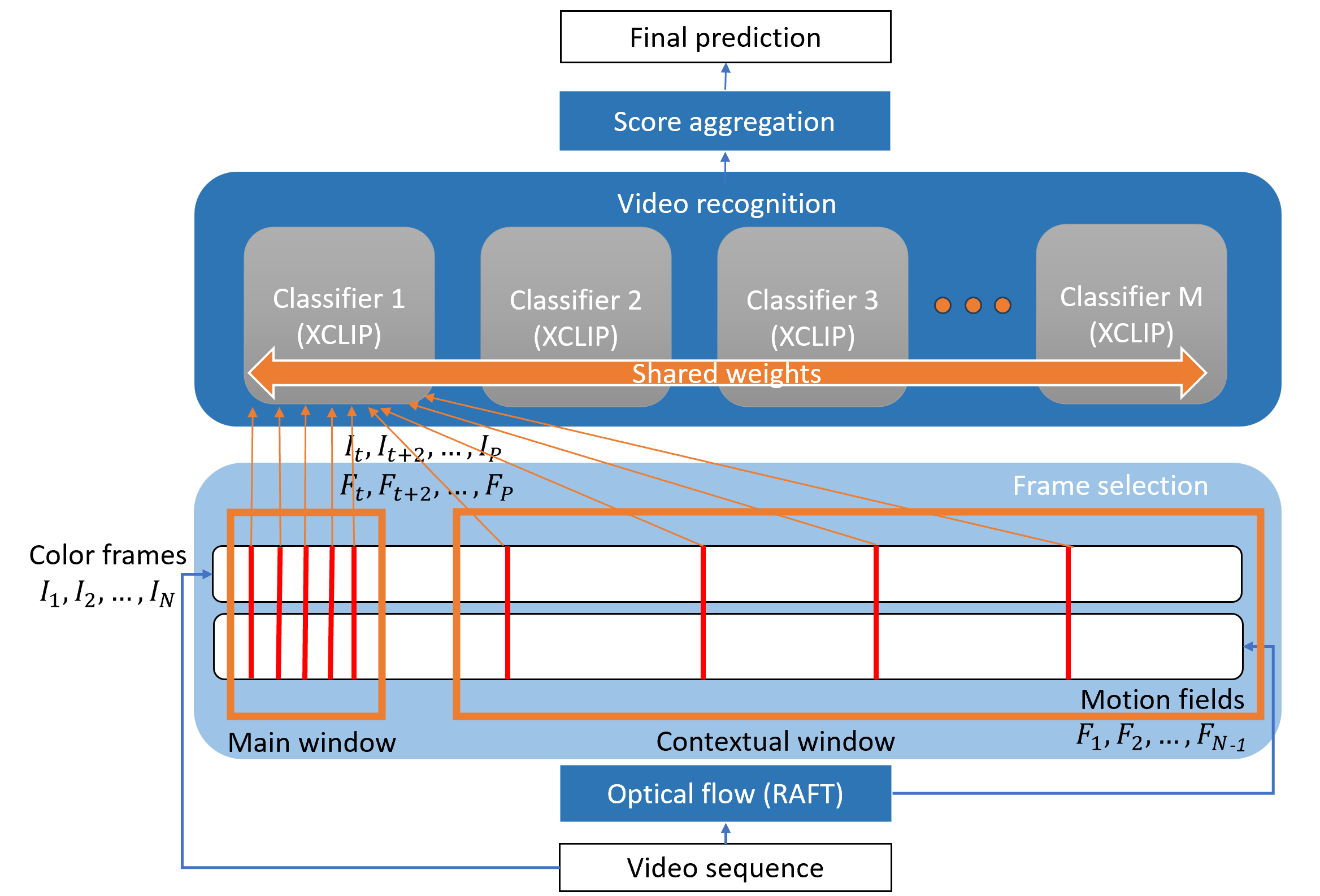}
    \caption{ Overview of the AnimalMotionCLIP system with an illustrative example of the frame selection process. The system processes color frames and motion fields from a video sequence, using a combination of main and contextual windows. Video recognition is performed by multiple XCLIP classifiers with shared weights, followed by score aggregation to generate the final prediction.
    The details are explained in \cref{sec:method}.  
    }
    \label{fig:overviewAnimalMotionCLIP}
\end{figure}

\begin{figure*}
\centering
\includegraphics[width= 0.8\textwidth]{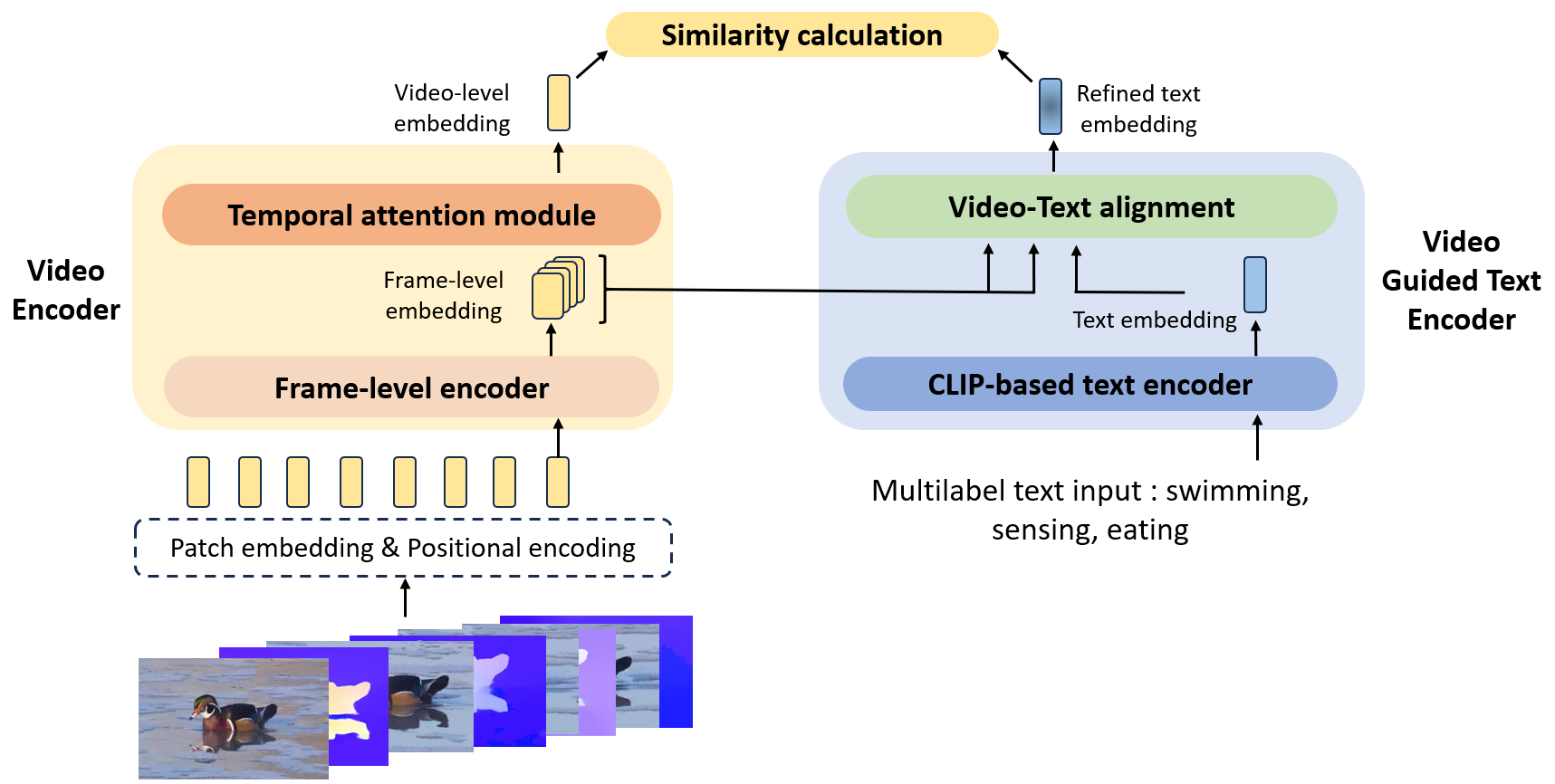}
\caption{Overview of the architecture of the classifiers in the Video Recognition module. It comprises a Video Encoder (left) and a Video Guided Text Encoder (right). 
}
\label{fig:detailedAnimalMotionCLIP}
\end{figure*}


The proposed AnimalMotionCLIP extends and adapts the visual-language model CLIP to video action recognition. As shown in \cref{fig:overviewAnimalMotionCLIP}, AnimalMotionCLIP integrates three main modules: Optical Flow, Video Recognition, and Score Aggregation. The input video $I=[I_1, I_2, ..., I_N]$ is processed by the Optical Flow module, obtaining the corresponding dense motion vector fields $F=[F_1, F_2, ..., F_{N-1}]$ using the RAFT algorithm \cite{teed2020raft}, where $F_i$ is a dense motion vector field computed between frames $I_i$ and $I_{i+1}$. RAFT has been chosen because of its high accuracy in near-motion boundaries and precision for small displacements, which is essential for capturing nuanced motion patterns in animal behaviors. Then, the video frames $I$ and the motion vector fields $F$ are processed by the Video Recognition module, which is formed by a set of $M$ classifiers based on XCLIP \cite{ni2022expanding} that share the same weights but process different video temporal content. The video content analyzed by every XCLIP classifier depends on two windows (the orange rectangles in \cref{fig:overviewAnimalMotionCLIP}): the main and the contextual ones. The main window, $W_m$, of a classifier is focused on a specific small part of the whole video, in such a way that the main windows of the other classifiers do not overlap with it and that all the main windows cover the entire video. Three frame selection schemes are proposed for the main and contextual windows. For the dense scheme, all the frames are evenly drawn only from the main window. For the semi-dense scheme, a selection of \(\frac{P}{2} \) video frames are drawn evenly for the main window (indicated by the red lines in \cref{fig:overviewAnimalMotionCLIP}). The contextual window $W_c$ covers the complementary frames of $W_m$, from which other \( \frac{P}{2} \) video frames are extracted. Finally, for the sparse scheme, no distinction between $W_m$ and $W_c$ is made, and frames are (random) quasi-evenly drawn from the whole video sequence, allowing slightly different frame selections for each classifier to acquire a more diverse context. Additionally, the corresponding motion vector fields are drawn for every sampling scheme and interleaved with the video frames as $IF = [I_{t_{1}}, F_{t_{1}}, I_{t_{2}}, F_{t_{2}}, ...]$. Thus, each XCLIP classifier processes different video frames and motion vector fields. The output of the Video Recognition module is a set of $M$ scoring vectors, $S=[S_1,..., S_M]$, one per classifier, indicating the probabilities of the actions/behaviors considered for different video temporal content. Finally, the Aggregation module, first, computes the average of all the scoring vectors $S_M$, and then thresholds it to select the most probable behaviors in the video. Notice that the system can predict multiple behaviors per video. 

\paragraph{Video Recognition module}

Every classifier of the Video Recognition module is based on the XCLIP architecture, which has been augmented to include video and optical flow information. The architecture comprises a Video Encoder and a Video-Guided Text Encoder. 
The Video Encoder computes an embedding containing a highly semantic representation of the whole video and the underlying motion using a combination of color frames and motion vector fields. The Video Guided Text Encoder computes an embedding for every word or text describing a behavior. The classifier is trained using the concept of contrastive learning\cite{radford2021learning} to produce similar embeddings for a video and its corresponding text (indicating one or multiple behaviors), while forcing the dissimilarity between non-related texts and videos. For this purpose, a sigmoid activation function per label and the cross-entropy loss function are used to enable multi-label (multi-class) classification. In inference, a K-Nearest Neighbor strategy is applied to infer the behavior in a video by comparing the resulting video embedding with a set of text embeddings (everyone describing the potential behaviors of interest).

The Video Encoder can be further divided into three submodules. The first one, Patch embedding and Positional encoding, divides each frame into $N$ non-overlapping patches of size \(P \times P \) and projects every patch onto an embedding of dimension $D$. Positional encoding is applied to these patch embeddings to retain spatial information. Subsequently, this sequence of patch embeddings is delivered to the Frame-level Encoder submodule, which uses a Multi-Head Self-Attention Mechanism (MHSA) architecture to compute a set of image embeddings capturing high-level semantic details for each frame. These two first submodules are similar to the architecture of a Vision-based Transformer \cite{dosovitskiy2021image}, but extend the concept to process multiple frames instead of a single image. Finally, the Temporal Attention submodule aggregates these frame-level embeddings over time. Using a standard Transformer architecture, it applies attention mechanisms across the sequence of frame embeddings to produce a single video embedding. This final embedding encapsulates both the content and the motion within the video.


The Video Guided Text Encoder is composed of the Text Encoder and the Video-Text alignment submodules.
The first component leverages a pre-trained CLIP-based text encoder to process the input multilabel text. Each word undergoes token embedding and positional encoding to create a highly semantic text embedding.
Following this, the Video-Text alignment submodule refines these text embeddings under the guidance of the video-level embedding, enhancing the capture of semantic correlations between the two modalities. This strategy substitutes the manual designed fixed prompt template used in other approaches for the input of the text encoder and provides dynamic adaptability to varying video content and contexts.

\section{Experiments}
\label{sec:experiments}

To evaluate the effectiveness of the proposed AnimalMotionCLIP, we conducted experiments on the Animal Kingdom dataset \cite{ng2022animal} to address the problem of general recognition of animal behavior. 
\begin{figure}
    \centering
    \includegraphics[width=1\linewidth]{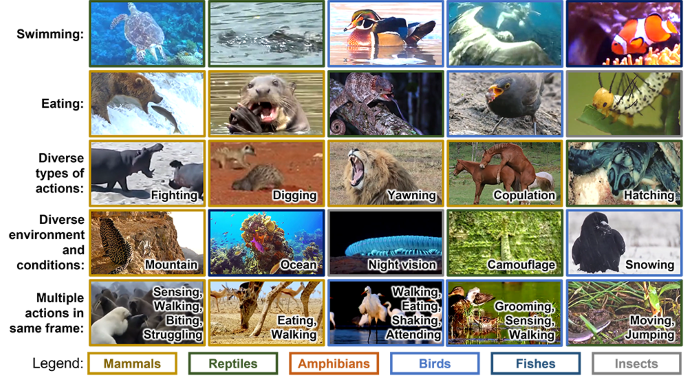}
    \caption{Example of the Animal Kingdom dataset with different animal species in different environments and exhibiting diverse behaviors.}
    \label{fig:enter-label}
\end{figure}
Following the guidelines of the Animal Kingdom dataset, the mean average precision (mAP) \cite{10.1145/2808194.2809481} is used as the evaluation metric, given by 

\begin{equation}
\begin{aligned}
                \text{mAP} =  \frac{1}{n} \sum_n (R_n - R_{n-1}) P_n, 
\end{aligned}
\label{eq:mAP}
\end{equation}

where $R_n$ and $P_n$ are the precision and recall at the threshold $n$. 

The authors of the Animal Kingdom dataset \cite{ng2022animal} presented results based on the convolutional neural networks: I3D\cite{carreira2018quo}, Slowfast \cite{feichtenhofer2019slowfast}, and X3D \cite{feichtenhofer2020x3d}.
Among these models, X3D achieves the highest mean average precision (mAP), progressively expanding a compact 2D image classification architecture into 3D. However, its performance remains relatively low compared to the results achieved by our proposed framework (see Table \ref{tab:comparison}). 
Regarding the state-of-the-art frameworks based on visual language models, two concurrent works are mostly connected to ours. The Category-CLIP\cite{10.1145/3581783.3612551} and MSQNet \cite{Mondal_2023}. 
Compared to these methods, our proposed AnimalMotionCLIP model demonstrates significant advances in animal behavior analysis (see Table \ref{tab:comparison}). Category-CLIP, which uses category-specific prompting techniques, enhances the CLIP model by incorporating an additional branch to predict animal classes, achieving a mean average precision (mAP) of 55.36\%. Similarly, MSQNet achieves an mAP of 55.59\% by leveraging the generalizability of the visual language model for the classification task.
In contrast, our AnimalMotionCLIP model surpasses both and achieves a significantly higher mAP of 74.63\% without relying on the computationally expensive pre-training strategies on additional Kinetic-400 dataset \cite{kay2017kinetics}, designed for human action recognition. By pre-training on this dataset, the MSQNet learns a set of features related to actions, which are then transferred to animal behavior recognition. 
 Even more, the proposed strategy still overcomes the accuracy of MSQNet + pre-training.

 \begin{table}
     \centering
     \begin{tabular}{cc}
         \toprule
            \textbf{Method}    & \textbf{mAP(\%)}  \\
        \midrule
                        I3D\cite{carreira2018quo}   & 16.48   \\
                        SlowFast\cite{feichtenhofer2019slowfast}  & 20.46 \\
                        X3D \cite{feichtenhofer2020x3d}  & 25.25 \\
                        Category-CLIP\cite{10.1145/3581783.3612551} & 55.36  \\
                        MSQNet  \cite{Mondal_2023} & 55.59   \\ 
                        MSQNet + pre-training \cite{Mondal_2023} & 73.10  \\ 
                        \textbf{AnimalMotionCLIP (ours)} & \textbf{74.63}  \\
    
        \bottomrule

     \end{tabular}
     \caption{Performance comparison with the state-of-the-art models on the Animal Kingdom dataset. The best results are in bold. }
     \label{tab:comparison}
 \end{table}

\paragraph{Ablation study} We first evaluated the most effective frame selection method using the RGB modality. As shown in \cref{tab:ablation_sampling} dense, semi-dense, and sparse sampling strategies were compared, yielding mAP scores of 68\%, 68.49\%, and 73.93\%, respectively. Dense sampling, which focuses on detailed segments, was found to be insufficient. Semi-dense sampling showed slight improvements, suggesting the need for more adaptable frame selection. The sparse sampling method, which uses quasi-random selection across videos, proved to be the most effective, capturing diverse animal behaviors and interactions more successfully. Subsequently, for the most effective sampling strategy (sparse one), we conducted a detailed analysis of the impact of motion information on behavior recognition performance. As shown in \cref{tab:ablation_modality}, using only RGB frames resulted in a mAP of 73.93\%, while using only optical flow frames yielded a slightly lower mAP of 65.1\%. However, combining RGB and optical flow frames led to an improved mAP of 74.63\%, demonstrating the complementary nature of both modalities and underscoring the importance of integrating visual and motion information for accurate behavior recognition.

 \begin{table}
     \centering
     \begin{tabular}{cc}
         \toprule
            \textbf{Modality}    & \textbf{mAP(\%)}  \\
        \midrule
            Dense &   68\\
            Semi-dense & 68.49 \\
            \textbf{Sparse} &  \textbf{73.93} \\
         
        \bottomrule

     \end{tabular}
     \caption{Performance comparison of temporal sampling strategies. The best results are in bold. }
     \label{tab:ablation_sampling}
 \end{table}

 \begin{table}
     \centering
     \begin{tabular}{cc}
         \toprule
            \textbf{Modality}    & \textbf{mAP(\%)}  \\
        \midrule
            RGB only &   73.93\\
            Optical flow only  & 65.1 \\
            \textbf{RGB + Optical flow} &  \textbf{74.63} \\
         
        \bottomrule

     \end{tabular}
     \caption{Performance comparison across different modalities. The best results are in bold. }
     \label{tab:ablation_modality}
 \end{table}

\paragraph{Error analysis} In the evaluation of AnimalMotionCLIP for animal behavior recognition, we observed several sources of inaccuracies that impacted the performance of the proposed framework.

One significant issue arises from the overlapping characteristics of an action. For instance, when the true label is `jumping', the model predicts a combination of `Keeping still' and `moving'. This error is likely due to the transitional aspect of jumping, in which an animal may be motionless before and after the jump, causing the model to distinguish these segments rather than the jump itself. In addition, the accuracy of the video annotations is sometimes arguable, since the ground-truth labels are not fully representative of the depicted behaviors. For example, a video showing a frog, first keeping still, and then jumping is annotated only as `jumping', ignoring the initial stationary phase. Furthermore, for the `sensing' class, the proposed framework sometimes predicts a combination of actions such as `attending', `Keeping still', and `sensing'. This suggests that the model can recognize related subactions, but struggles to consolidate them into a single, cohesive action label. Another challenge is the complex nature of certain actions, which are often composed of multiple subactions. For example, grooming behavior in animals can include various actions such as licking or scratching themselves. Our model tends to select a series of actions to describe this category, resulting in predictions like `attending', `eating', `keeping still', and `moving'" for a single grooming behavior. This can lead to discrepancies between the predicted and true labels.

Overall, these errors highlight the need for more refined annotation practices and enhanced model capabilities to better capture and distinguish between complex and composite behaviors in animal behavior recognition.
\section{Conclusion}
\label{sec:conclusion}

In this paper, we introduced AnimalMotionCLIP for animal behavior analysis, adapting visual language models to this specialized domain. Our approach addresses two critical challenges: integrating motion information and devising an effective temporal modeling scheme. By combining video frames and optical flow within the CLIP framework, AnimalMotionCLIP accurately correlates textual descriptions with observed animal behaviors. We compared various temporal modeling schemes, finding that sparse sampling was most effective for capturing diverse behaviors. Leveraging pre-trained visual language models, our framework ensures efficiency and domain-specific relevance without extensive pre-training. Experiments on the Animal Kingdom dataset show that AnimalMotionCLIP outperforms state-of-the-art approaches, demonstrating superior performance in recognizing fine temporal actions essential for animal behavior analysis.

\paragraph{Acknowledgement}
This research was funded by the grant TED2021-130225A-I00 (ANEMONA) funded by MCIN/AEI/10.13039/501100011033 and by the “European Union NextGenerationEU/PRTR”.

This work has also been partially supported by the Ministerio de Ciencia e Innovación (AEI/FEDER) of the Spanish Government under project PID2020-115132RB (SARAOS).

{
    \small
    \bibliographystyle{ieeenat_fullname}
    \bibliography{main}
}


\end{document}